\documentclass{ecai} 

\usepackage{tikz}
\usetikzlibrary{positioning,fit}

\usepackage{latexsym}
\usepackage{amssymb}
\usepackage{amsmath}
\usepackage{amsthm}
\usepackage{booktabs}
\usepackage{enumitem}
\usepackage{graphicx}
\usepackage{color}

\definecolor{obright}{HTML}{EFF2F1}
\definecolor{odark}{HTML}{283044}
\definecolor{oblue}{HTML}{6B9AC4}
\definecolor{ored}{HTML}{C33C54}
\definecolor{opurple}{HTML}{4A306D}

\newcommand{\BibTeX}{B\kern-.05em{\sc i\kern-.025em b}\kern-.08em\TeX}

\begin{document}


\begin{frontmatter}


\paperid{123} 


\title{Automated Explanation Selection for Scientific Discovery}


\author{\fnms{Ashlin}~\snm{Iser}\orcid{0000-0003-2904-232X}\thanks{Email: ashlin.iser@kit.edu}}

\address{Karlsruhe Institute of Technology, Germany}


\begin{abstract}
Automated reasoning is a key technology in the young but rapidly growing field of Explainable Artificial Intelligence (XAI).
Explanability helps build trust in artificial intelligence systems beyond their mere predictive accuracy and robustness.
In this position paper, we propose a cycle of scientific discovery that combines machine learning with automated reasoning for the generation and the selection of explanations.
We present a taxonomy of explanation selection problems that draws on insights from sociology and cognitive science.
These selection criteria subsume existing notions and extend them with new properties.
\end{abstract}

\end{frontmatter}

\section{Introduction}

The trustworthiness of artificial intelligence~(AI)~systems, including a sound rationale for their deployment, is paramount in safety-critical and high-risk applications such as critical infrastructure operations, healthcare diagnostics and medical devices, law enforcement and criminal justice, biometric identification and categorization of natural persons, or autonomous vehicles such as self-driving cars or drones~\cite{Rudin:2019:IntrinsicInterpretability}.
There are also legal reasons to want intelligible AI, such as the ``European Artificial Intelligence Act''~\cite{EU:2021:Regulation}. 
Discrimination based on gender, race, or origin is a significant challenge in many areas of contemporary society.
Discrimination can be exacerbated by the use of AI systems that are trained on data sets that reflect the same biases that exist in the wider world.
This becomes a major concern when AI systems are used to make decisions that \emph{affect the life of people}.

Despite their problem-solving capabilities, machine learning (ML) models often suffer from opacity, making their inner workings difficult or impossible for humans to understand.
Explainability contributes to building trust in AI~systems beyond their pure predictive accuracy and robustness~\cite{Weld:2019:IntelligibleAI}. 
Efficient automated reasoning enables exact approaches to the explainability of ML~models, and is thus a key technology of the young but rapidly growing field of explainable artificial intelligence~(XAI)~\cite{JMS:2023:LogicBasedExplainability}.
Formal methods facilitate the early detection of flaws in ML~models and the validation of ML~models by checking for their coherence with prior knowledge and expectations.
This facilitates the identification and prevention of flawed decisions and discriminatory practices through the use of AI~systems.

Since the training of an ML~model is an inductive process that exploits correlations in a set of observations, i.e.,~training data, the epistemological status of such learned abstractions is subject to the \emph{problem of induction}~\cite{Lange:2011:ProblemOfInduction}.
This means that while inductive inferences, such as the rules manifested in ML~models, are likely to be true based on the given data, they are not guaranteed to be true.
In contrast, deductive reasoning is characterized by certainty and precision: the conclusions are guaranteed to be true if the premise is true.
This implies that if the premise is an accurate representation of an ML~model, the conclusions are accurate if the model is accurate.
Conversely, if the conclusions are erroneous, the model is erroneous.
Therefore, any form of formal explanation extracted from an ML~model can merely serve as a hypothesis and can potentially be falsified by new observations. 

\begin{figure*}[htbp]
	\centering
	\begin{tikzpicture}[line width=1pt, draw=odark, >=latex, minimum width=7em]
		\node (data) {Data};
		\node (model) [left=10em of data] {Model};
		\node (encoding) [above=2em of model] {Encoding};
		\node (explanation) [right=10em of encoding] {Explanations};
		\node (hypothesis) [right=2em of explanation] {Hypothesis};
		\node (experiment) [below=2em of hypothesis] {Experiment};

		\draw[->] (data) -- node(ml)[below]{\footnotesize Machine Learning} (model);
		\draw[->] (encoding) -- node(ar)[below]{\footnotesize Automated Reasoning} (explanation);
		\draw[->] (hypothesis) -- (experiment);

		\node (induction) [fit=(data) (ml) (model), draw=oblue, dashed, label={[color=oblue,xshift=-3em]below left:{Induction}}, minimum height=2.5em] {};
		\node (deduction) [fit=(encoding) (ar) (explanation), draw=ored, dashed, label={[color=ored,xshift=-3em]above left:{Deduction}}, minimum height=2.5em] {};
		\node (abduction) [fit=(hypothesis) (explanation), draw=opurple, dashed, label={[color=opurple,xshift=-1em]above:{Explanation Selection}}, minimum height=3em] {};

		\draw[->] (model) -- (encoding);
		\draw[->] (explanation) -- (hypothesis);
		\draw[->] (experiment) -- (data);
	\end{tikzpicture}
\caption{Using explainable artificial intelligence to aid scientific discovery.}
\label{fig:xai4science}
\end{figure*}
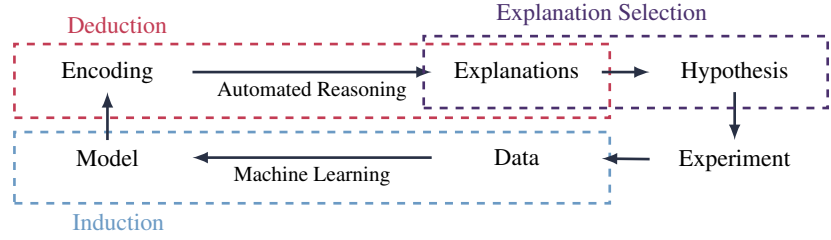

In this paper, we propose a machine-assisted cycle of scientific discovery which is illustrated in Figure~\ref{fig:xai4science}.
Our model combines machine learning with automated reasoning for the automated selection of explanations which can be used to formulate hypotheses that can be tested by experiments.
The blue box represents the inductive phase of the cycle, in which a model is learned from existing training data.
The red box represents the deductive phase, where the learned model is encoded into a formal language and automated reasoning is used to deduce explanations for the model.
The purple box represents the process of explanation selection which researchers can compare with prior knowledge and even use to formulate hypotheses.
These hypotheses can then be tested by further experiments, which in turn produce new data, thereby closing the cycle.

Modern automated reasoning systems are capable of producing certified deductive conclusions that can be checked by formally verified software.
This capability enables the reliable automatic analysis of ML models, including verification of ML models and extraction of knowledge from ML models to assist humans in learning from AI systems.
The present position paper will focus on this latter aspect.

We begin with a general discussion of the suitability of automated reasoning methods for generating explanations for machine learning models in Section~\ref{sec:method}.
Then, in Section~\ref{sec:properties}, we introduce a taxonomy of explanation selection problems that subsumes existing notions and extends them with new properties.
In particular, we discuss desirable properties of explanations drawn from the social sciences and how these can be used to guide automated explanation selection.
Finally, we discuss how the evaluation of explanations is simplified by the guarantees provided by automated reasoning methods in Section~\ref{sec:evaluation}, and then conclude with a summary of our findings.

\section{Automated Reasoning for Explainability}
\label{sec:method}

In recent years, considerable progress has been made in the field of automated reasoning, particularly with regard to algorithmic solutions to the propositional satisfiability problem (SAT)~\cite{Iser:SC2020}.
Automated reasoning and the SAT problem are closely related, with SAT solvers often being a central component in automated reasoning systems. 
Even though the complexity of SAT is challenging, the ubiquity of NP-hard problems has stimulated research to the extent that modern SAT solvers efficiently solve many SAT instances arising in practice.
In addition, modern SAT solvers are suitable for use in safety-critical and high-risk applications because they produce proofs that are given as efficiently checkable sequences of rigorous mathematical reasoning steps~\cite{Biere:2021:Handbook:Proofs}.

The genericity of the specification of the SAT problem is a key factor for their wide range of applications, including the verification of hardware 
and software, 
cryptanalysis, 
planning, 
and many more. 
Advances in SAT solving often lead to advances in related reasoning problems and their optimization variants, such as Maximum Satisfiability (MaxSAT) or Pseudo-Boolean Satisfiability (PBS) and Optimization (PBO). 
Such advances are also closely related to problems higher up in the polynomial hierarchy, such as Quantified Boolean Formulas (QBF). 
SAT solvers are also used in the backend of reasoning systems that support richer formal languages, such as SAT Modulo Theory (SMT) solvers~\cite{Biere:2021:Handbook}.

There are three main reasons why SAT-based methods are particularly well suited for this task.
First, recent advances in SAT solving show that SAT-based methods have the potential to efficiently solve the instances of the NP-hard explanation selection problems addressed in this paper. 
Second, SAT-based methods are particularly reliable because they produce proofs that can be efficiently checked for correctness by verified software. 
Finally, these methods are generic, allowing flexible adaptation to problem variants by modifying the encoding of the problem instance rather than the algorithm.

\subsection*{Efficiency and Trustworthiness}

Conflict-driven Clause Learning (CDCL) is the most popular SAT algorithm, because its implementations are extremely efficient in practice. 
Essential to its success are efficient data structures, implementations, and heuristics for its main pruning techniques, which are unit propagation and clause learning~\cite{Biere:2021:Handbook:CDCL}.

While solutions to the SAT problem by definition produce certificates of satisfiability where possible, state-of-the-art CDCL solvers also produce certificates of unsatisfiability that can be verified in time polynomial in the length of the proof.
Assuming ${\mathsf{NP} \neq \mathsf{co-NP}}$, the length of a polynomial time checkable proof is necessarily worst-case exponential in the size of the input formula.
But proof checking is feasible in practice, since the length of the proof correlates with the solving time, i.e., if we could generate it, we can check it~\cite{Biere:2021:Handbook:Proofs}.
Certificate checking is a distinct process from certificate creation and can be accomplished with formally verified software, thereby enhancing the dependability of the solver outcomes~\cite{Tan:2023:PropagationRedundancy}.
This is particularly crucial when it comes to verifying the safety properties of systems for high-risk applications.

The study of proof complexity provides the theoretical background for understanding some of the practical advances and limitations of logic solvers, in particular with respect to resolution-inspired clause learning in modern SAT solvers and the emergence of new heuristics for exploiting more powerful proof systems in SAT solving. 
CDCL solver runtimes are subject to the same domain-specific exponential lower bounds as proof-lengths in the general resolution proof-system they simulate~\cite{Biere:2021:Handbook:ProofComplexity}.
The study of more powerful proof systems, such as extended resolution, has a long tradition, and recent implementations of new practical heuristics manage to exploit them.
This can be achieved by the addition of propagation redundant clauses~\cite{Reeves:2023:Prelearn}, or symmetry-breaking clauses~\cite{Devriendt:2016:BreakId}, or variable definitions~\cite{Haberlandt:2023:SBVA} that help break lower bounds on the length of unsatisfiability proofs is a way to take advantage of more powerful proof systems.

While the SAT solver with the highest average performance may not be optimal in all scenarios, the use of a portfolio of SAT solvers can significantly improve the expected overall performance, indicating that the selection of a solver should be based on the domain to be solved~\cite{Bach:2022:kPortfolios}.
Recent advances have also built on the capabilities of parallel SAT solvers, with the most efficient being clause-sharing portfolios of CDCL solvers. 
The massively parallel SAT solver Mallob builds on a portfolio of some of the most successful sequential SAT solvers and employs a unique scalable clause-sharing strategy~\cite{Schreiber:2023:Thesis}.

The problem of enumerating all models of a satisfiable formula (AllSAT) has been addressed by a variety of approaches that can be broadly classified into two categories: blocking approaches, which are based on incremental SAT solving using minimized blocking clauses~\cite{Yu:2014:AllSAT}, and non-blocking approaches, which ensure exclusive enumeration through chronological backtracking~\cite{Grumberg:2004:MemoryEfficientAllSAT}.
A recent approach to AllSAT outperforms existing approaches by avoiding no-good learning through the combination of chronological backtracking and implicant shrinking~\cite{Spallitta:2023:EnumeratingDisjointModels}.

The propositional optimization problem (MaxSAT), is defined as the problem of finding an assignment that maximizes the number of satisfied clauses.
This problem has two notable variants: a partial variant, in which a subset of its clauses must be satisfied, and a weighted variant, in which the objective is to maximize a sum of clause weights. 
The MaxSAT algorithms that remain relevant today include model-improving approaches, also known as Linear SAT UNSAT (LSU), core-guided approaches (CG), and implicit hitting set approaches (IHS).
These algorithms essentially solve a sequence of SAT instances with an incremental CDCL SAT solver~\cite{Biere:2021:Handbook:MaxSAT}.
LSU algorithms solve a sequence of satisfiable instances by adding bound-tightening constraints until the instance is UNSAT.
Conversely, CG algorithms solve a sequence of unsatisfiable instances by relaxing the instance in each iteration until the instance is SAT.
Similar to core-guided approaches, IHS algorithms solve a sequence of unsatisfiable instances, but tighten the lower bound by computing a minimum-cost hitting set of the accumulated unsatisfiable cores.
Recent advances in MaxSAT solvers have been made in multiobjective optimization~\cite{Jabs:2023:MaxSATMultiObjective} and in incremental MaxSAT solving~\cite{Niskanen:2021:IncrementalMaxSAT}.

\subsection*{Formal Explanation Approaches}

In naturally interpretable models such as decision trees, evaluation traces can serve as direct explanations of prediction results~\cite{Rudin:2019:IntrinsicInterpretability}.
In practice, such models often have limited predictive power because they can only generate step-like decision boundaries, are mostly restricted to categorical data, and are prone to overfitting.
Moreover, according to recent research in formal explanability, explanations by interpretable models are generally not concise and exhibit redundancy~\cite{Izza:2022:ExplanationRedundancy}.

Heuristic explanation approaches sample the input space around the instance to be explained and observe the effect on the output space. 
As a result, these approaches can be applied to any class of prediction models.
Due to their probabilistic nature, such model-agnostic approaches do not provide strict guarantees of correctness or minimality of explanations.
A prominent example is LIME, a sampling-based approach that locally approximates a given model by an intrinsically interpretable model, such as a decision tree~\cite{Ribeiro:2016:LIME}.
However, heuristic explainers are not guaranteed to be correct, and their explanations can be misleading.
A formal analysis of the quality of the well-known Shapley values has shown that they can make misleading predictions about the importance of features~\cite{Huang:2024:ShapFail}. 
Anchors is an approach that computes sufficient conditions along with an estimate of their accuracy for a given prediction. 
Also for Anchors, a formal analysis has shown that it gives unrealistically high accuracy estimates of explanations~\cite{Narodytska:2019:AssessingSATBased}.

In contrast, explanations by formal methods are derived from an exact formal language encoding of the learned model and a formalization of the desired properties of the sought explanations.
The guarantees of formal approaches are mathematically exact, as they are based on sound deduction systems~\cite{Ignatiev:2020:Towards}. 
In recent years, the emerging formal XAI community has focused on the notions of abductive explanation (AXp) and contrastive explanation (CXp).
Here, an AXp, also known as prime implicant explanation (PI-explanation) or minimally sufficient reason, denotes a minimal set of features sufficient to guarantee a given prediction.
In contrast, a CXp, also known as minimal required change, denotes a minimal set of features sufficient to change a prediction~\cite{Ignatiev:2019:AbductionBased}. 
There are weak and strong variants in the literature to distinguish \emph{subset minimality} from \emph{cardinality minimality} for both AXps and CXps.
The computation of both AXps and CXps for tree ensembles is an NP-hard problem~\cite{Izza:2021:RFs}. 

Nevertheless, efficient formal explainers for tree ensembles have been developed, one based on recent SMT and one based on recent MaxSAT approaches.
In terms of runtime, both exact approaches outperform even the heuristic explainer Anchor~\cite{Ignatiev:2022:TreeEnsemblesMaxSAT}.
An important property of AXps and CXps is their duality, i.e.,~contrastive explanations are minimal hitting sets of abductive explanations and vice versa~\cite{Ignatiev:2020:c2a}.
Thus, AXp/CXp enumeration is closely related to the well-studied enumeration problems of Minimum Unsatisfiable Subsets (MUS) and Minimum Correction Sets (MCS) in formal logic which is $\Sigma^P_2$-hard~\cite{JMS:2020:InconsistentF}.
The state-of-the-art formal explanation enumeration algorithm for decision lists is one that exploits this duality, analogous to the generic MARCO algorithm for MUS/MCS enumeration~\cite{Ignatiev:2021:DLs,Liffiton:2016:FastMUS}.

Ensemble learning is a popular approach to improve the accuracy of ML models by combining multiple weak models into a single strong model~\cite{Hastie:2009:ESL}.
Tree ensembles, such as random forests and gradient boosting machines, are well-known examples of ensemble learning methods in which the weak models are decision trees.
Tree ensembles achieve high model accuracy with a minimal number of hyperparameters, and their generalization performance is robust even in high-dimensional feature spaces due to randomization in the training process and the combination of many decision trees with limited depth.
Although tree ensembles are formed from a large number of intrinsically interpretable decision trees, they themselves are \emph{not} intrinsically interpretable.

Studying the formal explainability of tree ensembles is a relevant direction of research because tree ensembles are a common class of ML models used in many practical applications, such as diagnosis~\cite{Yang:2023:MedicalApplications}, or marketing~\cite{Prabadevi:2023:CustomerChurning}.
Moreover, they can be used to explain the predictions of any black-box or otherwise hard-to-codify prediction model, such as a large neural network, as is done in Asteryx, which combines sampling-based explanation with formal explanation.
Asteryx uses sampling to locally approximate a given black-box model by a random forest trained on the samples drawn.
This random forest then serves as a surrogate model for the subsequent computation of exact formal explanations~\cite{Boumazouza:2021:Asteryx}.

\section{Desirable Properties of Explanations}
\label{sec:properties}

We find that the problem of explanation selection has been understudied, resulting in a literature that contains redundant and overlapping notions, some of which even have misleading names.
To alleviate these problems, we have reviewed the literature on relevant notions of explanation in the social sciences and have identified a set of desirable properties that can be used to guide the selection of explanations.

There is an extensive body of research from philosophy, psychology, sociology and cognitive science on the nature of explanations and what constitutes a good, useful or acceptable explanation~\cite{Miller:2019:XAISocialSciences}.
On the nature of explanations, we first note that an explanation is an answer to a ``why?'' question.
While the concrete question ``Why P?'' can be extended to ``Why P instead of not P?'', in informal communication a contrast case Q is given by the context, so that implicitly it is almost always a question of the form ``Why P instead of Q?''~\cite{Pearl:2018:BookOfWhy}.
Contrast cases, also known as foils, are key to providing the context for an explanation, and different contrast cases lead to different explanations~\cite{Chin-Parker:2017:ContrastiveConstraints}.
An effect can have multiple necessary and sufficient causes but full causal chains are often too large to comprehend~\cite{Hilton:2016:SocialAttributionAndExplanation}.
The contrast between \emph{the fact and the foil} is the primary criterion for selecting an explanation which highlights the differences between the contrasted cases~\cite{Lipton:1990:ContrastiveExplanation,Hilton:1996:MentalModels}.

Moreover, \emph{necessity and sufficiency} are strong criteria for selecting explanations~\cite{Lipton:1990:ContrastiveExplanation}.
\emph{Minimality and generality} are other desirable properties of explanations, as people tend to prefer shorter explanations and those that apply to many situations rather than just one particular situation~\cite{Thagard:1989:ExplanatoryCoherence,Woodward:2006:SensitiveAndInsensitiveCausation}.
Explanation selection can also be based on \emph{anomaly}, i.e.,~people tend to select unusual events as explanations~\cite{Hilton:1986:AbnormalConditionsFocusModel}.
In contrast, the most probable explanations are often assumed to be known~\cite{Hilton:1996:MentalModels}.
Explanations based on statistical generalization require an additional explanation of the \emph{causes for the statistical generalization}~\cite{Josephson:1994:AbductiveInference}.

Selection of useful explanations effectively reduces the exponential number of explanations to a digestible set.
This involves a systematic formalization of desirable properties, thereby extending and subsuming existing notions of explanation selection.
The diverse (and sometimes multiple) objectives can be then formulated as (multi-objective) MaxSAT problems.
This also allows us to formulate objective thresholds that can be used to reduce the number of explanations for enumeration approaches.

Necessity, sufficiency, and minimality have been studied in the context of formal explanation selection for machine learning models~\cite{Ignatiev:2019:AbductionBased}.
For minimality, we can formulate both weak and strong versions of some of those desirable properties to disambiguate subset-minimality and cardinality-minimality.
In addition, contrast cases can be specified in each of the resulting explanation selection problems.
Some desirable properties require reasoning about an underlying data set, which poses additional challenges for formal encoding of models.
This particularly affects \emph{generality}, i.e.,~conditions explaining the largest number of data points, and \emph{anomaly}, i.e.,~conditions explaining the smallest number of data points.

\subsection*{Relationship with Theories of Causality}

Even though explanation and causality denote two different things, explanation can be associated with the dependence theories of causality, which include Hume's theory of regularity~\cite{Hume:2000:Enquiry7}, later refined by Lewis, who defined causality as a matter of counterfactuals~\cite{Lewis:1973:Causation}.
This concept is closely related to explanation selection, but in this context we use counterfactuals to isolate desired explanations from the set of possible explanations. 

\subsection*{Relationship with Theories of Abduction}

Explanation selection is also closely related to the notion of abduction, which was introduced by Peirce as a third mode of reasoning in addition to deduction and induction.
Deduction is the process of deriving a result from a known fact and a rule. Induction, on the other hand, is the process of generalizing a rule from a set of observed facts and results. Abduction, on the other hand, is the process of proposing a fact to explain a known result using a known rule~\cite{Magnani:2023:HAC}.

There is a large body of work on abductive reasoning in the context of the philosophy of logic and scientific discovery~\cite{Magnani:2023:HAC}.
Peirce developed a theory of science in which abduction is the step in which a hypothesis is generated, i.e., a fact that explains an observation is suggested by scientific intuition, then deduction is used to evaluate the plausibility of the hypothesis, and then induction is used to provide further evidence for the hypothesis through experimentation.

In AI research, interpretations that capture abduction as a form of ``Inference to the Best Explanation'' (IBE) are common~\cite{Josephson:1994:AbductiveInference}.
In contrast, Hintikka's notion of abduction is embedded in an interrogative model of inquiry
~\cite{Hintikka:1999:LogicOfDiscovery} and understood as the introduction of new explanatory hypotheses, as opposed to IBE's reasoning with a fixed hypothesis space~\cite{Hintikka:1998:Abduction}.
This view is closer to Peirce's notion of abduction, emphasizing hypothesis formation over selection among given alternatives; whereas IBE reduces to deduction over a fixed set of hypotheses with additional preference criteria.

\section{Evaluation of Explanations}
\label{sec:evaluation}

The authors of a recent comprehensive survey of evaluation methods in Explainable Artificial Intelligence identified a total of \emph{twelve evaluation criteria} for the explanations generated by XAI tools~\cite{Nauta:2023:Evaluating}.
In the following, we argue that many of these criteria can be fully satisfied a priori by formal reasoning methods.

The criteria \emph{correctness}, \emph{completeness}, and \emph{consistency}, are fully satisfied by SAT-based formal reasoning methods, since they produce certified results that can be validated with verified software.
Exact formal approaches can also explicitly maximize the \emph{continuity} and \emph{compactness} of the explanations by incorporating them as objective functions in the encoding of the explanation selection problem, while at the same time preserving the correctness and completeness of the encoding.

However, \emph{coherence} with background knowledge, beliefs, etc.~is an inappropriate criterion for evaluating explanations in the context of formal reasoning.
This is because, due to the exactness of the formal method, any incoherence of explanations necessarily follows from an incoherence of the model.
In the context of heuristic explainers, which are not guaranteed to be exact, it may make sense to check whether the resulting explanations are consistent with what is expected from the background knowledge.
In the formal setting, however, if the exact explanation is not consistent with the background knowledge, then the model is not consistent with the background knowledge.
Thus, coherence is not a criterion for evaluating the explanation, but rather a criterion for evaluating the model.
Formal verification of ML~models is a natural complementary application to explanation selection.
Verification of ML~models can be achieved by encoding expectations as additional constraints in the model encoding to allow proof by contradiction.

Since SAT-based formal reasoning produces exact and verified results, the \emph{confidence} in the generated explanations is generally high. 
However, the whole process still depends on the correctness and completeness of the SAT encoding of the model, as illustrated by the edge connecting the red and blue boxes in Figure~\ref{fig:xai4science}.
This involves parsing the data structures of the ML~model $M$ and generating a formal language representation $F$ that preserves the semantics of~$M$.
This means that for any input $I$ and output $O$, the formula $F$ has a model if and only if $M(I) = O$, provided that the input and output vectors $I$ and $O$ are represented by variables of $F$.
This encoding phase must be verified by additional means, such as formal proofs of the correctness and completeness of the encoding and, for testing practical implementations, automated consistency tests.

The \emph{contrastivity} criterion evaluates the possibility of specifying contrast cases.
Formal encodings of ML~models should therefore explicitly allow the specification of contrast cases in the encoding of the explanation selection problem.

Other evaluation criteria such as \emph{controllability}, \emph{covariate complexity}, \emph{composition}, and \emph{user satisfaction} are not directly addressed by the formal reasoning methods themselves, but can be achieved through the design of an appropriate user interface for specfication of user constraints, and for presentation and visualization of explanations and explanation statistics.

\section{Conclusion}

We proposed a cycle of scientific discovery that combines machine learning with automated reasoning for generating and selecting explanations. 
Our findings indicate that automated reasoning methods are well-suited for generating explanations for machine learning models and can guide their selection. 
We also introduced a taxonomy of explanation selection problems, drawing on insights from sociology and cognitive science, which subsumes existing notions and extends them with new properties.
We showed that evaluation is greatly simplified through the guarantees provided by automated reasoning methods. 
Overall, applying automated reasoning to explanation selection for machine learning models represents a promising direction for future research.



\begin{ack}
I would like to express my gratitude to Prof. Arnaud Durand (Univ. Paris Cité) and Prof. Christine Roussat (Univ. Clermont Auvergne) for their invaluable feedback, which they provided through discussions or comments on earlier drafts of this paper.
\end{ack}



\bibliography{ecai}

\end{document}